\documentclass[10pt,twocolumn,letterpaper]{article}

\usepackage{wacv}
\usepackage{times}
\usepackage{epsfig}
\usepackage{graphicx}
\usepackage{amsmath}
\usepackage{amssymb}
\usepackage{comment}
\usepackage[ruled,vlined,linesnumbered]{algorithm2e}
\usepackage{multirow}
\newcommand{\sceme}{\texttt{SCEME}\xspace}
\usepackage[margin=0pt,font=small,labelfont=bf]{caption}
\newcommand{\squishlist}{
   \begin{list}{$\bullet$}
    { \setlength{\itemsep}{0pt}      \setlength{\parsep}{3pt}
      \setlength{\topsep}{3pt}       \setlength{\partopsep}{0pt}
      \setlength{\leftmargin}{1.0em} \setlength{\labelwidth}{1em}
      \setlength{\labelsep}{0.5em} } }

\newcommand{\squishend}{
    \end{list}  }

\newcommand{\PP}[1]{\noindent{\bf #1}}

\def\wacvPaperID{1106} 

\wacvfinalcopy 

\ifwacvfinal
\fi

\ifwacvfinal
\usepackage[breaklinks=true,bookmarks=false]{hyperref}
\else
\usepackage[pagebackref=true,breaklinks=true,colorlinks,bookmarks=false]{hyperref}
\fi

\begin{document}

\title{ADC:  \underline{A}dversarial attacks against object \underline{D}etection that evade \underline{C}ontext consistency checks}

\author{Mingjun Yin\thanks{equal contribution} \and
Shasha Li\footnotemark[1]  \and 
Chengyu Song\and 
M. Salman Asif \and 
Amit K. Roy-Chowdhury \and 
Srikanth V. Krishnamurthy \\
University of California, Riverside\\
\tt\small \{myin013,sli057\}@ucr.edu, csong@cs.ucr.edu, \{sasif,amitrc\}@ect.ucr.edu, krish@cs.ucr.edu}

\maketitle
\thispagestyle{empty}

\begin{abstract}
Deep Neural Networks (DNNs) have been shown to be vulnerable to adversarial examples, which are slightly perturbed input images which lead DNNs to make wrong predictions. To protect from such
examples, various defense strategies have been proposed. A very recent defense strategy 
for detecting adversarial examples, that has been shown to be robust to current attacks, 
is to check for intrinsic context consistencies in the input 
data, where context refers to various relationships (e.g., object-to-object co-occurrence relationships) in images. In this paper, we show that even context consistency checks can be brittle 
to properly crafted adversarial examples and to the best of our knowledge, we are the first to do so.
Specifically,
we propose an adaptive framework to generate examples that subvert such defenses, namely, \underline{A}dversarial attacks against object \underline{D}etection that evade \underline{C}ontext consistency checks (ADC). In ADC, we formulate a joint optimization problem which has two attack goals, viz., (i) fooling the object detector and (ii) evading the context consistency check system, at the same time. Experiments on both PASCAL VOC and MS COCO datasets show that examples generated with ADC fool the object detector with a success rate of over 85\% in most cases, and at the same time evade the recently proposed context consistency checks, with a 
``bypassing'' rate of over 80\% in most cases. Our results suggest that ``how to robustly model context and check its consistency,'' is still an open problem.

\end{abstract}

\section{Introduction}
\begin{figure*}
\centerline{\includegraphics[width=1.5\columnwidth,keepaspectratio]{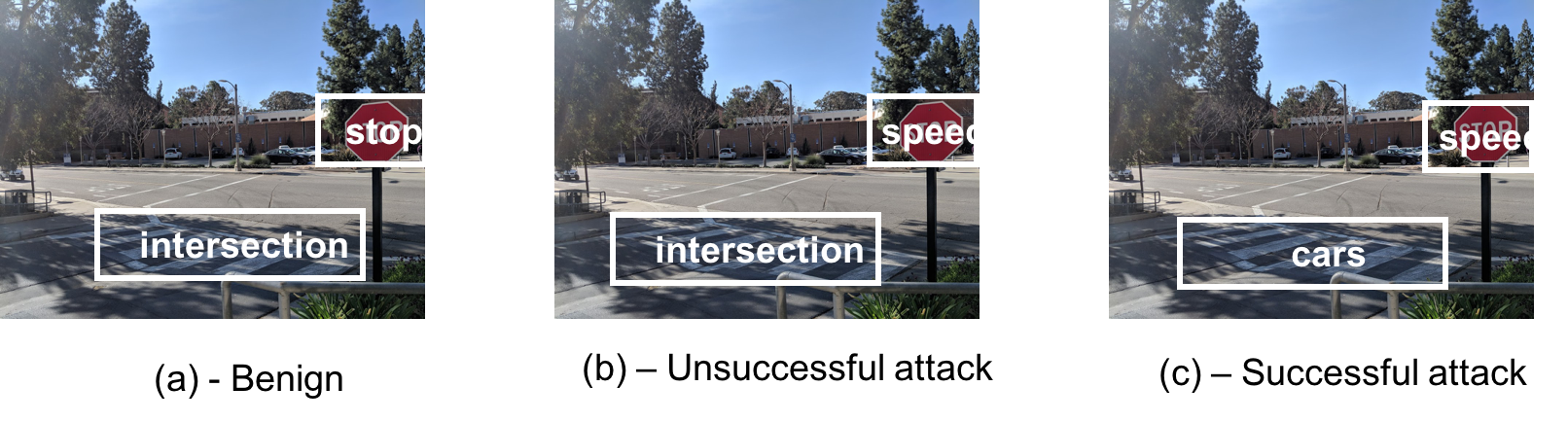}}
\caption{Motivation for our proposed approach. (a) A benign image where a detector 
detects the stop sign and intersection. (b) Only the stop sign is perturbed and detected as a speed limit; however, a context-aware adversarial defense like ~\cite{li2020connecting} is able to easily detect that a speed limit sign does not occur at an intersection, and thus the attack fails. (c)
Our goal is to not only perturb the stop sign into a speed limit sign, but also perturb other parts of the image so that it is contextually consistent, e.g., the intersection to objects that co-occur with the speed limit sign like other cars. Such an attack would be difficult to detect using methods like ~\cite{li2020connecting}. While we do not explicitly perturb the other objects, the proposed method will add perturbations such that the detections in the overall image become contextually consistent.}
\label{fig:motivation}
\end{figure*}
Deep Neural Networks (DNNs) have been shown to be highly expressive and have achieved state-of-the-art (SOTA) performance on a wide range of computer vision tasks, such as object detection and classification.
However, in 2014, Szegedy~\etal~\cite{szegedy2013intriguing} found that DNNs are vulnerable to carefully crafted and usually, visually inconspicuous (to humans) perturbed inputs named adversarial examples, which lead DNNs to make incorrect predictions.
Since then, an arms race between the generation of adversarial example attacks and defenses 
to thwart them, has taken off. 
Researchers have proposed attacks and defenses in image and video classification~\cite{sharif2016accessorize,li2020measurement,li2018adversarial,wei2019sparse,xu2017feature,jia2019identifying,xiao2019advit,li2021adversarial}, 
and object detection~\cite{chen2018shapeshifter,song2018physical,zhao2019seeing,li2020connecting}.

Among existing defense mechanisms, checking intrinsic context consistencies within the input data has recently been showcased to be very effective, in various tasks. For example, spatial consistency has been used to detect adversarial attacks against semantic segmentation~\cite{xiao2018characterizing}; temporal consistency has been used to detect adversarial attacks against video classification~\cite{jia2019identifying,xiao2019advit}; object-object context along with other kinds of context has been used to detect adversarial examples against objection detection~\cite{li2020connecting}; audio-visual correlation has been used to detect adversarial examples against audio-visual speech recognition~\cite{ma2019detecting}.

While these context-consistency-based defenses are shown to be effective against SOTA attacks,
whether they can resist more advanced adaptive attacks remains unexplored.
We hypothesize that if the context is (or can be) extracted using a neural network and the consistency checks are also performed using neural networks, then adaptive attacks could be feasible 
via a more complex optimization formulation.
Specifically, we expect that it would be feasible to use existing optimization techniques to 
compute adversarial examples that can (a) fool the DNNs to make wrong prediction (e.g., misclassify an object) and (2) fool the consistency check modules thus bypassing the defense. This is shown diagrammatically using an example in Fig. \ref{fig:motivation}.

In this work, we conduct the {\em first} study on composing adaptive attacks against context-consistency-based defenses.
Using the very recent context-aware object detector proposed in~\cite{li2020connecting} 
as an exemplar of defense, 
we demonstrate the feasibility of attacking context-consistency based models.
We analyze two types of attack scenarios that are as follows.
In a white-box attack scenario, we assume attackers have full knowledge of the consistency check modules.
In our gray-box attack scenario, we assume attackers are aware of the consistency check mechanism (e.g., the neural network architecture), but have no access to either the parameters of the modules or the training data.

Specifically, we develop a framework that we call ADC for ``Adversarial attacks against object
Detection that evade Context consistency checks'' to develop white box based adversarial examples.
To generate these white-box adaptive attacks in ADC, we formulate a joint optimization problem where the loss function is composed of two parts: (a) the loss function of the DNN prediction, and (b) the loss function relating to the consistency check.
One particular challenge in solving the joint optimization problem is that in current context aware defenses, the context is extracted from the intermediate layers of the DNN, while the consistency checks are done with separate DNN module(s).
For instance, the context-aware object detector proposed in~\cite{li2020connecting} extracts context profiles from internal Gated Recurrent Units (GRUs) and checks for relationship inconsistencies using a bank of auto-encoders.
To solve this challenge, we propose a 3-step optimization technique.

With respect to the gray-box adaptive attack, our aim is to expand ADC to investigate the transferability of ``context-aware'' attacks.
Specifically, since we do not have full access to the consistency check modules, we train a surrogate defense module and then solve the joint optimization problem with the surrogate module.
Then, we test the generated attacks with the gray-box model.

The main contributions of our work are as follows:
\squishlist
	\item To the best of our knowledge we are the first to investigate adaptive adversarial attack against consistency check based defenses. 
	\item We develop a framework, ADC, for generating practically viable adaptive adversarial examples against context consistency based defenses. To generate adversarial examples using ADC, we formulate a joint optimization problem, and find the approximate solution efficiently by solving three sub-optimization problems in a pipeline.
	\item We conduct extensive experiments on the large-scale PASCAL VOC and MS COCO dataset. Our method yields very high fooling rates against the object detection system while simultaneously, evading the context consistency check with  high success rates, for three typical attack goals~\cite{chen2018shapeshifter,song2018physical,zhao2019seeing}: mis-categorization, hiding and appearing attacks. 
\squishend

\section{Related Work}

To craft adversarial examples, early methods~\cite{szegedy2013intriguing,goodfellow2014explaining,kurakin2016adversarial} either take one step to apply perturbations to the input along the sign of gradient of the designed misclassification loss function, or take multiple small steps iteratively while adjusting the direction after each step. Some notable works that considered both attacks and defenses include  ~\cite{papernot2016distillation,carlini2017towards,guo2017countering,xie2017mitigating,dhillon2018stochastic,athalye2018obfuscated}.

One promising defense strategy explored recently, is to leverage violations of intrinsic {\bf consistencies} that are expected to exist in the input data (i.e., context) to detect adversarial examples. Specifically, context relates to 
relationships
that are typical in the training samples; adversarial examples are expected to cause violations in such relationships.
Xiao \etal \cite{xiao2018characterizing} show via an empirical study that the surrounding spatial context information in semantic segmentation has different characteristics for benign and adversarial examples, and proposed a spatial consistency-based adversarial detection method.
Jia \etal \cite{jia2019identifying} define an exception frame as one whose prediction label differs from the labels of adjacent frames and showed that exception frames are more likely in adversarial video inputs than in benign video streams. 
Xiao \etal \cite{xiao2019advit} proposed to use temporal consistency to detect adversarial examples in video.
Ma \etal \cite{ma2019detecting} found that the correlation between audio and video in adversarial examples is lower than benign examples due to the added perturbations. 
Yin \etal \cite{yin2021explpoiting} abstracted a pseudo-language to describe the scene and utilizing natural language processing technique to measure the overall context consistency for a given image.
Wang \etal \cite{wang2021multiexpert} proposed a context consistency check mechanism to detect adversarial attacks against ReID systems.

In a very recent work, Li \etal \cite{li2020connecting} construct the context profile, which captures four types of relationships among region proposals (i.e., spatial context, object-object context, object-background context, and object-scene context) in the object detection task to detect adversarial perturbations. Based on the observation that the context profile of each object category is usually unique, they use a bank of auto-encoders to learn the benign context profile distribution for each category and detect adversarial attacks as context consistency violations (i.e., high reconstruction error of auto-encoders) during testing. Experiments show that the detection ROC-AUC is over 0.95 for various adversarial attacks. Although their method has been very successful with most attacks that are currently prevalent, in this paper, we show that it is possible to design context-aware attacks that can defeat such systems.
To the best of our knowledge, we are the first to propose adaptive attacks to evade consistency check based defenses in object detection. 

\section{The ADC Framework}
\begin{figure*}
\centerline{\includegraphics[width=2
\columnwidth]{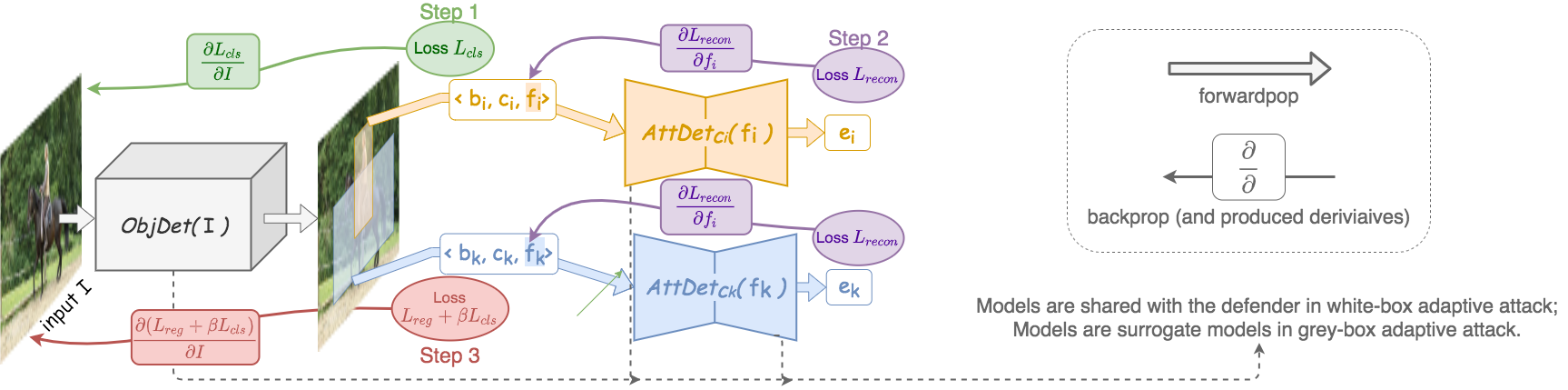}}
\caption{Our proposed ADC framework. There are two attack goals in ADC: a) fooling the object detector; b) bypassing the context-inconsistency checks. The first goal is achieved with the optimization problem defined in step 1 which perturbs the input image to make the object detector output the target classification labels. The second goal is achieved with the two optimization problems defined in step 2 and step 3. In step 2, the context profile features $f_{i}$ are perturbed to make the attack detector (auto-encoders) output low anomaly scores. In step 3, the input image is further perturbed to output the target perturbed context profiles and the target classification labels. Note that in gray-box attack, the $ObjDet(\cdot)$ and $AttDet_{c}(\cdot)$ are trained separately by the attacker, 
and it is not the same model used by the defender.}
\label{fig:framework}
\end{figure*}

In this section, we describe how we design our ADC framework to generate adaptive attacks against object detection systems with a context-inconsistency-based adversarial example detection mechanism.
We first give a brief introduction of 
how a context-inconsistency-based attack detection mechanism like~\cite{li2020connecting} works. 
Then we describe the threat model in detail. Subsequently, 
we formulate an optimization problem to jointly fool the detector and evade 
the context consistency check (referred to as joint optimization) to generate our white-box adaptive attack and 
describe how we find approximate solutions to this optimization problem via a three-step strategy.
Lastly, we describe the strategy we use in ADC to realize a gray-box adaptive attack.

\subsection{Context-inconsistency-based attack detection}
\label{sec:sceme}
In this subsection, we illustrate how context-inconsistency-based attack detection mechanisms work
In general, these defense mechanisms extract some context profiles from the input data and check whether the extracted context profiles match some known distributions.
If a testing time context profile does not fit into a known distribution, then an anomaly is detected.
Such an anomaly could be introduced by previously unknown data or adversarial attacks.

Take the \sceme model proposed by Li~\etal~\cite{li2020connecting} as an example.
\sceme aims to defend the Faster RCNN~\cite{ren2015faster} object detector against adversarial examples.
Given an input image, Faster RCNN will output a set of region proposals, each of which is associated with bounding box information, and a category probability vector.
To detect adversarial attacks, \sceme extracts a context profile for each proposed region.
Each context profile is a vector of intermediate features, computed based on a fully connected context graph, where each node is a region proposal and edge weights are learned from the training data.
Because each category of objects is likely to have distinguished object-object relationships, \sceme uses an auto-encoder to learn the benign distribution of context profiles for each category.
To detect context inconsistencies, the context profile is input to the auto-encoder associated with the predicted category.
A high reconstruction error of the auto-encoder implies an abnormal context profile and thus, the corresponding region would be marked as containing adversarial perturbations.

\subsection{Problem definition and threat model}
In this subsection, we generalize and formulate the problem of attacking an object detector with context-inconsistency-based attack detection, such as the model proposed by Li~\etal~\cite{li2020connecting}.
We denote the input scene image as $I$, the object detection function as $\textit{ObjDet}(\cdot)$, and the detection results as $R_I = \textit{ObjDet}(I) = [r_1, r_2, \dots, r_N]$, where $N$ is the total number of proposed regions over $I$.
For simplicity of exposition, we will use $R$ instead of $R_I$ hereafter.
To detect adversarial attacks, the context-consistency checker extracts a context profile for each proposed region.
Therefore, for each region proposal, we have the predicted category label denoted as $c_i$, the bounding box coordinates denoted as $b_i$, and the context profile denoted as $f_i$, i.e., $r_i = < b_i, c_i, f_i>$.
If we use $C=[c_1, c_2, ..., c_N]$ to denote the predicted categories of all region proposals, $B=[b_1, b_2, \dots, b_N]$ to denote corresponding bounding boxes, and $F=[f_1, f_2, \dots, f_N]$ to denote all context profiles, then $R=<B,C,F>$.
Because each category of objects is likely to have distinguished object-object relationships, 
we denote the function of the attack detector (i.e., auto-encoder) associated with each category $c \in L$ ($L$ is the complete category set) as $\textit{AttDet}_{c}(\cdot)$.
For each region, we can further calculate the reconstruction error of a context profile as $e_i = \textit{AttDet}_{c_i}(f_i)$.


We aim to perform targeted attacks where the perturbed object is misclassified to an attacker desired category.
Compared to untargeted attacks, targeted attacks are harder and potentially provide more insights since subversion to certain target categories could be harder to bypass the context-consistency checks than others.
We consider two threat models in designing ADC.
\squishlist
	\item \textbf{White-box adaptive attack} where we assume attackers have full knowledge about the context-consistency check mechanism, i.e., $\textit{ObjDet}(\cdot)$ and $\textit{AttDet}_{c}(\cdot), \forall c \in L$ are known functions to attackers.
	\item \textbf{Gray-box adaptive attack} where we assume the attackers are aware of the context-consistency check mechanism, e.g., the definition of context profile and the use of the auto-encoder bank. However, they have {\em no} access to either the parameters of the models or the training data. In other words, $\textit{ObjDet}(\cdot)$ and $\textit{AttDet}_{c}(\cdot), \forall c\in C$ are unknown functions to attackers, but the attackers could estimate these functions based on their own training samples.
\squishend

\subsection{Joint optimization for white-box attack}
Our hypothesis is that if the context profile is extracted and checked using a neural network, and the parameters of the neural network are known (i.e., the white-box attack scenario), then we should be able to search for adversarial perturbations that can both fool the object detector and satisfy context-consistency constraints using joint optimization.
In this subsection, we describe how to realize our white-box adaptive attack.
Specifically, we will use the \sceme model as an example to show how we calculate context-aware perturbations that can both fool its object detector and bypass the inconsistency check applied by its auto-encoder bank.

\subsubsection{Formulation}

\PP{Fooling the object detector.}
To fool the object detector to misclassify the victim object into a target label, perturbations should be calculated such that the labels of the corresponding regions are flipped from $c_s$ to $c_t$, where $c_s$ denotes the ground truth label of the victim object, and $c_t$ denotes the target label assigned by the attacker. We use $C^*$ to denote the target labels for all the region proposals where ground truth labels are used for untargeted objects and label $c_t$ are used for the targeted object. We use $\Delta I_{\text{fool}}$ to denote the perturbation needed to fool the object detector. To provide imperceptibility of the perturbation, the $L_p$ norm (e.g. $L_1$, $L_2$, $L_{\inf}$ and etc.) of $\Delta I_{fool}$ is constrained to be lower than a
chosen threshold viz., $\tau_{\text{fool}}$.  
Therefore, the optimization problem can be formulated as in Eqn.~\ref{equ:cls}. The classification loss $L_{cls}$ is the cross-entropy loss between the predicted probability vector $P_{C}$ from the object detector and the target label $C^*$. Note that, in the optimization process we do not change the bounding box coordinates of the region proposals.
\begin{equation}
\label{equ:cls}
	\begin{aligned}
		& \underset{\Delta I_{\text{fool}}}{\text{minimize}}
		&& L_{cls}(P_C, C^*), P_C \gets \textit{ObjDet}(I+\Delta I_{\text{fool}})\\
		& \text{subject to}
		&& \| \Delta I_{\text{fool}} \|_{p} \leq \tau_{\text{fool}}
	\end{aligned}
\end{equation}

\PP{Bypassing the context-inconsistency checks.}
To bypass context-inconsistency checks, perturbations should be calculated such that the context profiles of all region proposals should have below threshold anomaly scores
(i.e., all auto-encoders in \sceme exhibit low reconstruction errors).
We use $\Delta I_{\text{bypass}}$ to denote the perturbation needed 
{\color{black} on the image pixels,} to bypass the context-inconsistency check.
The optimization problem can be formulated as in Eqn.\ref{equ:recon}.
The reconstruction loss $L_{recon}$ is the smooth $L_1$ loss as used in \cite{li2020connecting} between the original context profile and the context profile reconstructed with the auto-encoders {\color{black} and $\tau_{\text{bypass}}$ is
the threshold that bounds the perturbation}.
Note that the auto-encoder used for each context profile $f_i$ could be dynamically chosen according to the predicted category $c_i$.
\begin{equation}
\label{equ:recon}
	\begin{aligned}
		& \underset{\Delta I_{\text{bypass}}}{\text{minimize}}
		&& \sum_{f_i \in F} L_{recon}(f_i, \textit{AttDet}_{c_i}(f_i)), \\
		&
		&& F \gets \textit{ObjDet}(I+\Delta I_{\text{bypass}})\\
		& \text{subject to}
		&& \|\Delta I_{\text{bypass}}\|_{p} \leq \tau_{\text{bypass}}
	\end{aligned}
\end{equation}
\PP{Joint optimization formulation.}
If we denote the perturbations on the input image $I$ as $\Delta I$,
then the overall optimization problem formulation, that would yield a perturbation that fools the object detector, while at the same time evades the context-inconsistency checks is given by Eqn. \ref{equ:joint}. In this equation, \textcolor{black}{$\alpha$ is a parameter that is used to achieve a good trade-off between fooling
the object detector and evading the consistency check. $\Delta I$ is the total adversarial perturbation applied on the image, which combines the fooling and bypassing perturbations.} 
{\color{black} $\tau$ is the threshold which bounds the total perturbation
that can be applied (to ensure imperceptibility).}
\begin{equation}
\label{equ:joint}
	\begin{aligned}
		& \underset{\Delta I}{\text{minimize}}
		&& L_{cls}(P_C, C^*) + \alpha \sum_{f_i \in F} L_{recon}(f_i, \textit{AttDet}_{c_i}(f_i)), \\
		&
		&& P_{c}, F \gets \textit{ObjDet}(I+\Delta I)\\
		& \text{subject to}
		&& \|\Delta I\|_{p} \leq \tau
	\end{aligned}
\end{equation}
\subsubsection{Three-step strategy}
Unfortunately, the joint optimization problem in Eqn.~\ref{equ:joint} cannot be solved directly, for two reasons:
(1) the anomaly detector $\textit{AttDet}_{c}(\cdot)$ could be dynamically chosen according to the predicted category of each region proposal and
(2) the anomaly detection networks could be disconnected from the object detection network and in such cases (as in [13]), {\color{black} back-propagation over the pair of networks (object detection and anomaly detection) is infeasible.
}
To overcome these key challenges, we propose a three-step optimization to search for an approximate solution, as depicted in Algorithm~\ref{algo:attack}.


In step one \textcolor{black}{(Lines 1-4 in Algorithm~\ref{algo:attack})}, we use existing attack techniques like IFSGM~\cite{kurakin2016adversarial} to search for perturbations $\Delta I_{\text{fool}}$ that can cause the target object to be misclassified (i.e., solving Eqn. \ref{equ:cls}). $t$ is the iteration counter for IFSGM.
We denote the perturbed image calculated in step one as $I' = I + \Delta I_{\text{fool}}$. Similar to {\cite{carlini2017towards}, $\textit{NormProjection}(\cdot)$ is to project the perturbation back to the norm ball so that the $L_p$ norm of the perturbation is under the pre-defined threshold $\tau_{\text{fool}}$. For example, {\color{black} if the $L_{\inf}$ norm} is used, then $\textit{NormProjection}(x)$ is equal to $\tau_{\text{fool}}$ if $x >\tau_{\text{fool}}$, and is equal to $-\tau_{\text{fool}}$ if $x <-\tau_{\text{fool}}$\footnote{{\color{black} Note that appropriate thresholds are used when this function used at different places
in the algorithm.}}, and is equal to $x$ otherwise.

Next, we extract the context profiles $F$ (which consists of node features, and relationships between region proposals as edge features~\cite{li2020connecting}), from the perturbed image $I'$ \textcolor{black}{(Line 6 in Algorithm~\ref{algo:attack}) }.
In step two \textcolor{black}{(Lines 7-11 in Algorithm~\ref{algo:attack}) }, we search for perturbations $\Delta F$, such that the perturbed context profiles yield low reconstruction errors (i.e., solving Eqn.\ref{equ:recon_step1}).
\begin{equation}
\label{equ:recon_step1}
\begin{aligned}
& \underset{\Delta F}{\text{minimize}}
&& \sum_{f_i \in F} L_{recon}(f_i+\Delta f_i, \textit{AttDet}_{c_i}(f_i+\Delta f_i)) \\
& \text{subject to}
&& \|\Delta f_i\|_{p} \leq \tau_{F}, \forall \Delta f_i \in \Delta F 
\end{aligned}
\end{equation}
%
We denote the perturbed context profiles as $F^* = F + \Delta F$, and will 
use these as the target context profiles in step three.
Note that Eqn.\ref{equ:recon_step1} is different from Eqn. \ref{equ:recon}; in Eqn. \ref{equ:recon} the optimization is on the input image instead of extracted the context profiles.
Similarly, Eqn. \ref{equ:recon} involves both the object detector and the auto-encoder bank instead of only the auto-encoder bank in Eqn.\ref{equ:recon_step1}. 
In step three \textcolor{black}{(Lines 12-16 in Algorithm~\ref{algo:attack}) }, with the target context profiles $F^*$ and the target category labels $C^*$, we search for perturbations $\Delta I'_{\text{bypass}}$ so that the 
perturbed image with this additional perturbation i.e., $I'' = I' + \Delta I'_{\text{bypass}}$, will output both the target category labels $C^*$ and the target context profiles $F^*$ (i.e., solving Eqn \ref{equ:joint_step3}).
\begin{equation}
\label{equ:joint_step3}
	\begin{aligned}
		& \underset{\Delta I'_{\text{bypass}}}{\text{minimize}}
		&& \sum_{f_i \in F} L_{reg}(F, F^*) + \beta L_{cls}(P_C, C^*) \\
		&
		&& P_C,F \gets \textit{ObjDet}(I'+\Delta I'_{\text{bypass}})\\
		& \text{subject to}
		&& \|\Delta I_{\text{fool}} + \Delta I'_{\text{bypass}} \|_{p} \leq \tau 
	\end{aligned}
\end{equation}

\begin{algorithm}[t]
	\fontsize{9}{9}
	\selectfont
	\SetAlgoLined
	\SetKwInOut{Input}{Input}
	\SetKwInOut{Output}{Output}
	\Input{$I$, $C^*$, $\textit{ObjDet}(\cdot)$, $\textit{AttDet}_{c}(\cdot), \forall c\in L$}
	\Output{Adversarial example  $I'$}
	$I'$ = $I$\\
	\For {$t \gets 1$ to $T_1$}
	{
	$\Delta I_{\text{fool}} \gets \text{solve }  Equ.~\ref{equ:cls} \text{ with }I' \text{ and  } C^*$\\
	$I' = I + \textit{NormProjection}(I' + \Delta I_{\text{fool}} - I)$\\
	}
	$F \gets \textit{ObjDet}(I')$\\
	$F^* = F$\\
	\For {$t \gets 1$ to $T_2$}
	{
	$\Delta F \gets \text{solve }  Equ.~\ref{equ:recon_step1} \text{ with }F^*$\\
	$F^* = F + \textit{NormProjection}_{F}(F^* + \Delta F - F)$\\
	}
	$I'' = I'$\\
	\For {$t \gets 1$ to $T_3$}
	{
	$\Delta I'_{\text{bypass}} \gets \text{solve }  Equ.~\ref{equ:joint_step3} \text{ with }I', C^* \text{ and  } F^*$\\
	$I'' = I + \textit{NormProjection}(I'' + \Delta I'_{\text{bypass}} - I)$\\
	}
	
	\Return $I'$
	\caption{White-box Adaptive Attack}
	\label{algo:attack}
\end{algorithm}

\subsection{Leveraging transferability for gray-box attack}
The proposed framework can be extended to a gray-box setting. As noted in Fig. \ref{fig:framework}, in this setting, the attacker needs to train its own model for the context-aware object detection. We assume that the attacker has access to data similar to that used by the defender, i.e., data comes from the same distribution and label space; however, the samples used to train the models on the attacker's side is different from that on the defender's side. This makes the optimization problem mentioned above more challenging as the object detection and context profiles are expected to have higher uncertainty. This is analyzed experimentally in \S~\ref{sec:grey-box}.


\section{Experimental Analysis}
\label{sec:eval}
\begin{figure*}
\centerline{\includegraphics[width=1.8\columnwidth]{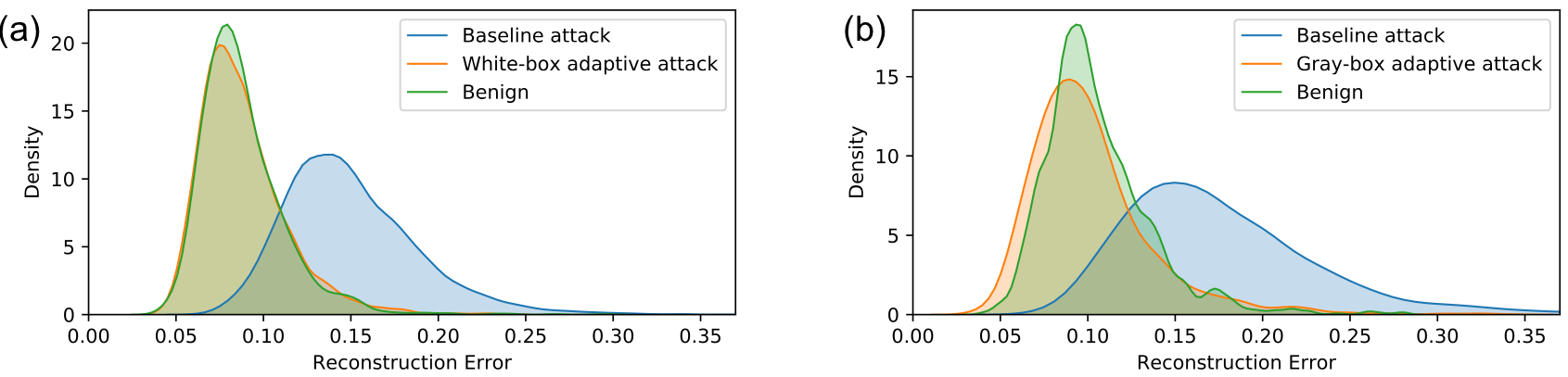}}
\caption{(a)The reconstruction error distributions of the context profiles from benign samples, adversarial samples generated by the baseline non-adaptive method, and adversarial sample generated by ADC method. The adversarial examples are generated in the white-box setting. We observe that ADC method is able to perturb the context profiles and mimic the benign context profile distributions.(b)The reconstruction error distributions of the context profiles from benign samples, adversarial samples generated by the baseline non-adaptive method in the white box setting, and adversarial sample generated by ADC method in the gray-box setting. We observe that when the attacker has no access to the oracle models, the context profile distribution leant with surrogate models is very similar to the ground truth benign context profile distribution. The context profiles transfer well across models, so as the perturbation on the context profiles.}
\label{fig:recon_dist}
\end{figure*}

\begin{table*}[!t]
\centering
\caption{Attack performance for three attack goals on the PASCAL VOC dataset.}
\label{tab:performance}
\small
\resizebox{14cm}{!}{
\begin{tabular}{c|ccc|ccc}
\hline
\multirow{2}{*}{Attack Method} & \multicolumn{3}{c|}{Fooling Rate}              & \multicolumn{3}{c}{Bypass Rate}                \\ \cline{2-7} 
                                & Mis-categorization & Hiding      & Appearing   & Mis-categorization & Hiding      & Appearing   \\ \hline
Non-adaptive attack (baseline) & 97.02\%            & 98.57\%     & 90.51\%     & 19.92\%             & 30.14\%     & 43.62\% \\ \hline
Adaptive attack (ours)         & 88.63\%            & 98.57\%     & 68.20\%      & 90.51\%             & 87.20\%     & 96.39\% \\ \hline
\end{tabular}}
\end{table*}

\begin{table*}[t]
\centering
\caption{Attack performance for three attack goals on the MS COCO dataset}
\label{tab:performance_COCO}
\small
\resizebox{14cm}{!}{
\begin{tabular}{c|ccc|ccc}
\hline
\multirow{2}{*}{Attack Method} & \multicolumn{3}{c|}{Fooling Rate}              & \multicolumn{3}{c}{Bypass Rate}                \\ \cline{2-7} 
                                & Mis-categorization & Hiding      & Appearing   & Mis-categorization & Hiding      & Appearing   \\ \hline
Non-adaptive attack (baseline) & 93.07\%            & 97.27\%     & 78.13\%     & 11.89\%        & 11.19\%         & 19.35\% \\ \hline
Adaptive attack (ours)         & 86.90\%            & 90.82\%     & 66.36\%     & 83.67\%        & 75.52\%         & 86.08\% \\ \hline
\end{tabular}}
\end{table*}

We conduct comprehensive experiments on two large-scale object detection datasets to evaluate attacks generated by ADC in both white-box and gray-box settings. Inspired by previous works~\cite{chen2018shapeshifter,song2018physical,zhao2019seeing}, we evaluate our approach with three attack goals:

\squishlist
\item \textit{Mis-categorization attack} where the object detector is to mis-categorize the perturbed object as belonging to a different category.

\item \textit{Hiding attack} where the object detector is to fail in recognizing the presence of the perturbed object, which happens when the confidence score is low or the object is recognized as background.

\item \textit{Appearing attack} where the object detector is to wrongly conclude that the perturbed background region contains an object of a desired category.
\squishend

For the mis-categorization attack and the appearing attack, we perform targeted attacks and randomly choose the target category labels, in our experiments.

We evaluate the performance of ADC with two metrics. Recall that we have two goals: (a) to fool the object detector to achieve the mis-categorization attack, the hiding attack, or 
the appearing attack; and (b) to bypass the consistency checks of the attack detectors (i.e., the auto-encoder bank).

\squishlist
\item \textit{Fooling rate} is the metric for the former goal, and
indicates how many attacks from all the tried ones, 
succeed in fooling the object detector.
 
\item \textit{Bypass rate} is the metric for the later goal, 
and quantifies how many attacks from those that fool the object detector, bypass the context consistency checks.
\squishend

An image is detected as natural/benign if the reconstruction errors of all the context profiles in the image, are lower than the threshold.
It is detected as containing adversarial perturbations, if the reconstruction error of any context profile in the image is higher than the threshold.
In other words, the maximum reconstruction error computed from the image (referred to as reconstruction error for simplicity hereon), is used to decide whether the image has violated the context consistency check or not.
To report the bypass rate, we need to fix the threshold.
The threshold for the reconstruction error is chosen so as to make the false positive rate equal to 0.1; here, the false positive rate is the probability that a benign image is wrongly detected as perturbed.  

\subsection{Implementation details}

\PP{Datasets.}
We use both PASCAL VOC~\cite{everingham2010pascal} and MS COCO~\cite{lin2014microsoft}. PASCAL VOC contains 20 object categories. Each image, on average, has 1.4 categories and 2.3 instances. \textit{VOC07trainval} and \textit{VOC12trainval} are used as the training set, and the the testing is carried out on \textit{VOC07test}. MS COCO contains 80 categories. Each image, on average, has 3.5 categories and 7.7 instances. \textit{coco14train} and \textit{coco14valminusminival} are used for training, and the test evaluations are carried out on \textit{coco14minival}. Note that COCO has few examples for certain categories. Similar to~\cite{li2020connecting}, we evaluate with  the 11 categories that have the largest numbers of extracted context profiles. 

\PP{Attack Implementations.}
To launch our adaptive attack, we add perturbations on the whole image, which is typical for digital attacks.
We will show the experimental results on adding perturbations to individual object regions in \S~\ref{sec:ablation}.
The hyper-parameters $T_1$, $T_2$, and $T_3$ are empirically chosen to be 10, 1, and 1 separately.
The step size for updating the perturbation on the input image is 1 and that for updating the perturbation on the context profile is 0.1.
$L_{inf}$ is used and the $\tau$ is set to be 10, and is the same as in previous works~\cite{moosavi2017universal,reddy2018nag,li2018adversarial}. $\tau_F$ is set to be 0.1. 
$\beta$ is set to be 1.

\begin{table*}[t]
\centering
\caption{The gray-box attack performance for three attack goals on the PASCAL VOC dataset}
\label{tab:performance_grey}
\vspace{-0.5em}
\small
\resizebox{14cm}{!}{
\begin{tabular}{c|ccc|ccc}
\hline
\multirow{2}{*}{Threat Model} & \multicolumn{3}{c|}{Fooling Rate}              & \multicolumn{3}{c}{Bypass Rate}                \\ \cline{2-7} 
                                & Mis-categorization & Hiding      & Appearing   & Mis-categorization & Hiding      & Appearing   \\ \hline
White-box         & 88.63\%            & 98.57\%     & 68.20\%      & 90.51\%             & 87.20\%     & 96.39\% \\ \hline
Gray-Box               & 88.05\%        & 85.33\%     & 68.09\%         & 92.34\%             & 86.57\%    & 96.38\% \\ \hline
\end{tabular}}
\end{table*}

\begin{table*}[t]
\centering
\caption{Attack performance when attacking only the target object region for three attack goals on the PASCAL VOC dataset}
\label{tab:performance_object}
\vspace{-0.5em}
\small
\resizebox{14cm}{!}{
\begin{tabular}{c|ccc|ccc}
\hline
\multirow{2}{*}{Attack Region} & \multicolumn{3}{c|}{Fooling Rate}              & \multicolumn{3}{c}{Bypass Rate}                \\ \cline{2-7} 
                                & Mis-categorization & Hiding      & Appearing   & Mis-categorization & Hiding      & Appearing   \\ \hline
 Whole Image                    & 88.63\%            & 98.57\%     & 68.20\%      & 90.51\%             & 87.20\%     & 96.39\% \\ \hline
Target Object         & 66.74\%        & 88.63\% & 34.54\% & 88.70\%        & 89.33\% & 95.15\% \\ \hline
\end{tabular}}
\end{table*}

\subsection{Evaluation of white-box attack performance}

\noindent \textbf{Baseline.} To understand the performance of the proposed context-ware adaptive attack method, we use a vanilla non-adaptive attack method as the baseline where the perturbation is calculated purely according to Eqn.~\ref{equ:cls} (i.e., simply aiming to fool the object detector without bypassing the context-consistency checks).

\noindent \textbf{Visualizing the reconstruction error.} We plot the reconstruction error calculated from benign images, perturbed images with the non-adaptive method, and perturbed images with the adaptive method in {\color{black} Fig.~\ref{fig:recon_dist}(a)}, with mis-categorization as the attack goal. We observe that the distribution of reconstruction errors of benign images differs from the distribution of perturbed images. Specifically, the reconstruction errors of the perturbed images are generally higher because context is violated in these cases. This is why, the context consistency check based detection works in~\cite{li2020connecting}. After we apply the perturbation with the proposed adaptive attack method, the reconstruction error of the perturbed images is low and the distribution is very similar to the that of the benign images. We also plot the distribution figure per target attack category in the supplementary material and we have consistent observations for all the categories. Therefore, we conclude qualitatively that the adaptive attack method is context-aware and offers promise in bypassing the context consistency check. 

\noindent \textbf{Adaptive attack performance.} To quantitively evaluate the adaptive attack performance, we report the fooling rate and the bypass rate on both PASCAL VOC and MS COCO in Tab.~\ref{tab:performance} and Tab.~\ref{tab:performance_COCO}.
We observe that ADC achieves a lower fooling rate compared to the non-adaptive baseline attack. For example the fooling rate of mis-categorization attack on the PASCAL VOC dataset decreases from 97.02\% to 88.63\%.
However, ADC significantly outperforms the baseline method in terms of the bypass rate.
For example, only 19.92\% of the baseline mis-categorization attacks evade the context-inconsistency based attack detection, while 90.51\% of the mis-categorization attacks launched by ADC evade this attack detection. 
We observe that on the MS COCO dataset, ADC improves the bypass rate from less than 20\% (with the baseline) to more than 75\% for all the three attack goals (mis-categorization, hiding and appearing), which corresponds to over a 55\% improvement.

\subsection{Evaluation of gray-box attack performance}
\label{sec:grey-box}

In gray-box attack, we use \textcolor{black}{\textit{VOC12trainval}} as the training set to train a surrogate system (object detector + auto-encoder bank) and use \textit{VOC07test} to train the target system's model. We generate perturbations by solving the optimization problem with the surrogate neural networks and the test it on the target system.
 
\noindent \textbf{Visualizing the reconstruction error.} We calculate the reconstruction error with the target system and plot the distributions in Fig.~\ref{fig:recon_dist}(b). The attack goal is also mis-categorization. We observe that although the perturbations are calculated with the surrogate system, they seem to transfer well onto the target system and the distribution of the reconstruction error of the perturbed images mimicked that of the benign images quite well. 

\noindent \textbf{Adaptive attack performance.} We report in Tab.~\ref{tab:performance_grey} the fooling rate and bypass rate of both white-box and gray-box adaptive attacks for all the three attack goals. We observe that compared to the white-box attack, the fooling rates of gray-box attack for all the three attack goals are slightly lower. However, we observe that, the bypass rates in gray-box setting are comparable to the white-box setting. This together with Fig.~\ref{fig:recon_dist}(b), implies that the context profile distribution learnt with surrogate models is  very similar to the ground truth benign context profile distribution. The context profiles transfer well across models, and so do the perturbations on the context profiles.

\subsection{Ablation study}
\label{sec:ablation}
\begin{figure}
\centerline{\includegraphics[width=0.8\columnwidth]{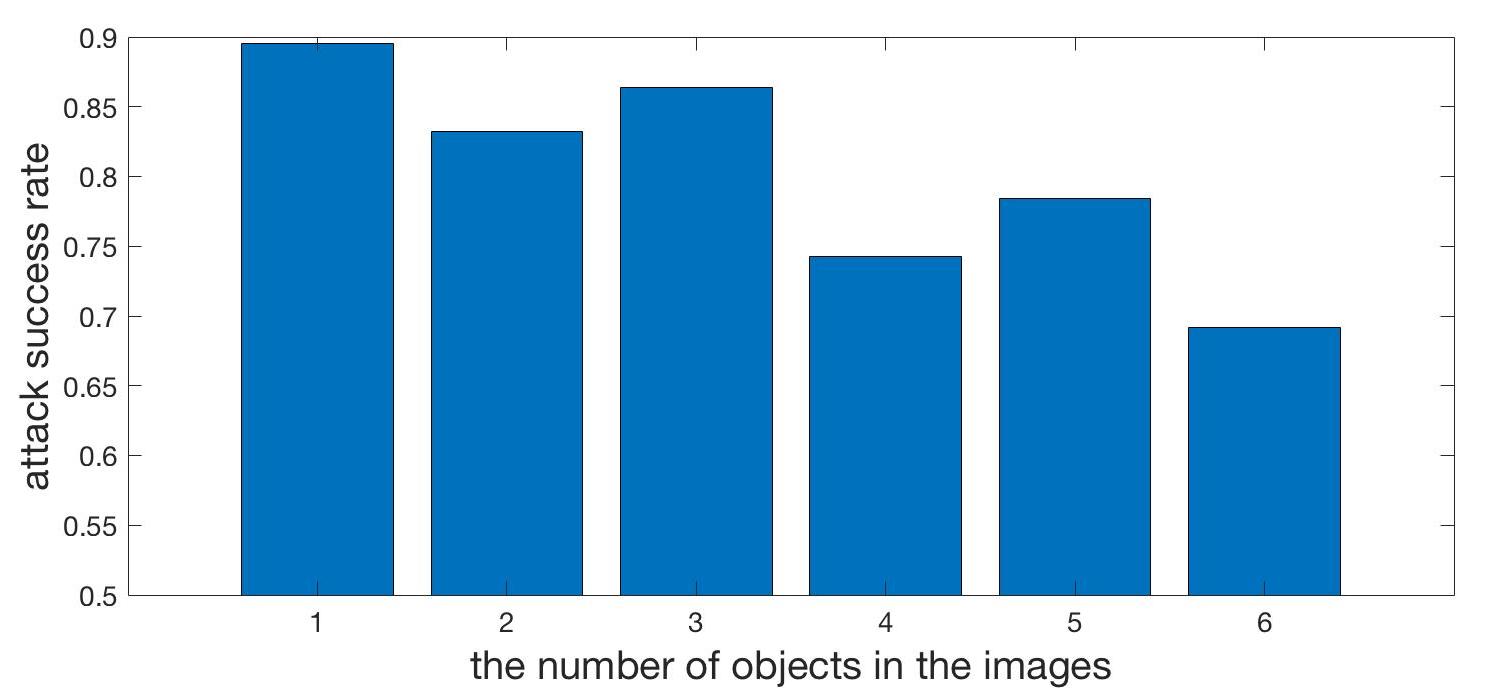}}
\caption{The attack success rates on MS COCO dataset for images with different number of objects. We observe that when more objects present, the attack success rates (fooling rate * bypass rate) tend to decrease. The reason could be with more objects, it is harder to perturb the target object and all the other objects in a context consistent way. }
\label{fig:num_obj}
\end{figure}

Next, we explore if we can bypass the context consistency check by perturbing only the target object region (defined by the ground-truth bounding box). By constraining the perturbation region, we have less power on perturbing the context profiles.
The results are shown in Tab.~\ref{tab:performance_object}. Compared to whole-image perturbations, although the single-object perturbations are able to retain comparable bypass rates, they lead to much lower fooling rates. The fooling rates of mis-categorization, appearing, and hiding attacks drop from 88.63\% to 66.74\%, 
from 68.20\% to 34.54\%, 
and from 98.57\% to 88.63\%, respectively.
This result is aligned with the expectation that (a) perturbing a single object is more likely to cause context violations in mis-categorization and appearing attacks and thus the attacks are less likely to succeed, and (b) hiding attacks are more likely to succeed because they generally do not cause context violations.
\begin{figure}[!h]
	\centering 
	\includegraphics[width=0.7\linewidth]{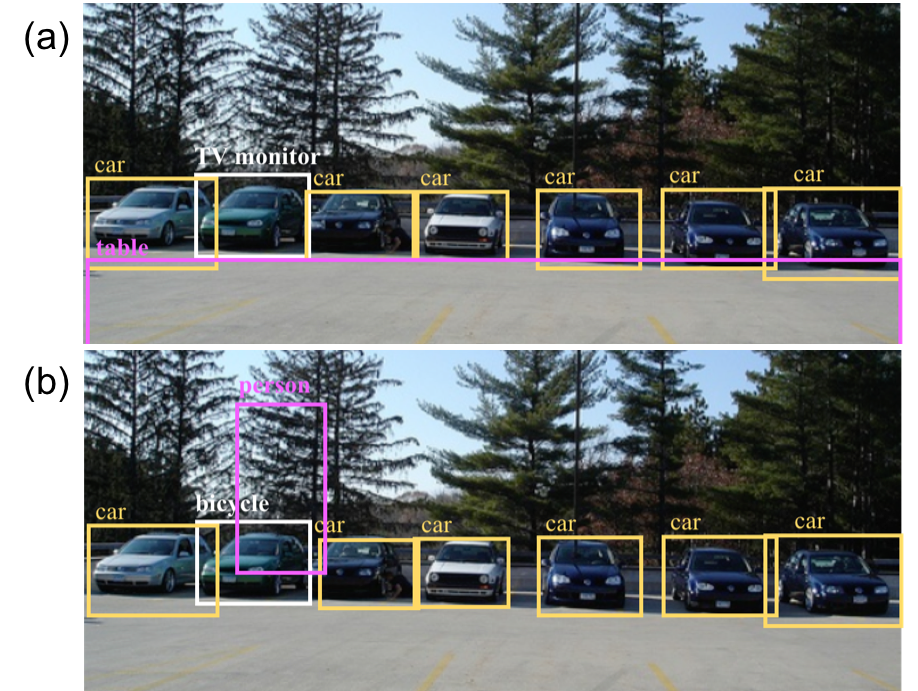}
	\caption{Two adversarial attack examples: (a) to misclassify the selected car instance into TV monitor and bypass the context-inconsistency check, ADC perturbed a background region into table; (b) to misclassify the selected car instance into bicycle and bypass the context-inconsistency check, ADC perturbed a background region into person.} 
	\label{fig:eg}
\end{figure}

\subsection{Analysis Study}
\label{sec:cases}
We explore what kind of attacks are harder to succeed compared to others. Specially, we test whether the number of objects in the images affect the attack performance. Here we define a new metric, attack success rate, which describes how many attacks out of all the tried ones succeed in {\em both} fooling and bypassing (i.e., captured by fooling rate multiplied by bypass rate). 
As shown in Fig.~\ref{fig:num_obj}, we observe that when more objects are present, the attack success rates tend to decrease.
We suspect the reason is that with more objects, it is harder to perturb the target object and all the other objects in a context-consistent way. 

We then dive deep into one adaptive attack example and see how the context is perturbed.
Fig.~\ref{fig:eg}(a) attack goal is to mis-categorize the selected car instance into a TV monitor.
After attacking with ADC, we observe that not only the car instance is misclassified to TV monitor, the parking lot is also misclassified as a table instance to make the context more consistent and thus help to bypass the consistency check.
Similarly, in Fig.~\ref{fig:eg}(b), when we try to mis-categorize the selected car instance into a bicycle with ADC, we observe that a person instance appears, overlapping the ``bicycle'' to make the context more consistent.

\section{Conclusions}

In this paper, we show that recent defense strategies that use
context consistency checks for detecting adversarial examples, can
be subverted by appropriately crafted attacks that jointly consider
the objectives of fooling the object detector and bypassing the consistency
checks. We develop a framework which we call ADC, to generate both adaptive white-box
and gray-box attacks, that are successful in this joint endeavor.
To the best of our knowledge, we are the first to show this possibility,
and our work highlights the inadequacies in current context models for
defending adversarial examples. Our experiments on both the PASCAL VOC and
MS COCO datasets show very high rates of both fooling the object detector (typically 
over 85 \%)
and evading the context consistency checks (typically over 80 \%). We believe that
future research on building
better context models, possibly tailored to adversarial example defense, may be needed
to truly make such approaches robust.

\noindent \textbf{Acknowledgments.}
{
This material is based upon work supported by the Defense Advanced Research Projects Agency (DARPA) under Agreement No. HR00112090096.
Approved for public release; distribution is unlimited.
}

{\small
\bibliographystyle{ieee_fullname}
\bibliography{egbib}
}
\appendix
\clearpage
\twocolumn[{%
 \centering
 \LARGE Supplementary Material for ``ADC: \underline{A}dversarial attacks against object \underline{D}etection that evade \underline{C}ontext consistency checks"\\[1.5em]
}]

\normalsize
In this supplementary material, we provide:
(a) the reconstruction error distribution plots of the context profiles for each object category,
(b) gray-box attack performance on the MS COCO dataset, and
(c) ablation study on the MS COCO dataset.

\section{Reconstruction error distribution}
As shown in Fig3.(a) in the main paper, the reconstruction error distribution of the context profiles from ADC perturbed images is very similar to that from benign images, and that is why ADC attack can evade the context-inconsistency based defense. In other words, the proposed ADC method achieves context-aware attacks to fool both the object detector and the attack detector.

Note that in the defense system, there is one auto-encoder for one category.
What is plotted in Fig3.(a) is the reconstruction error distribution for all the categories on the PASCAL VOC dataset.
We in this section present the reconstruction error distribution per object category.
Fig.~\ref{fig:density_per_class} shows the results.
As we can see, our previous observation holds for each category (subplot), i.e., the distribution of the ADC generated images mimics that of the benign images for each object category. This further proves that ADC can generate context-aware attacks that are able to bypass context-consistency checks.

\begin{figure*}
\centerline{\includegraphics[width=2
\columnwidth]{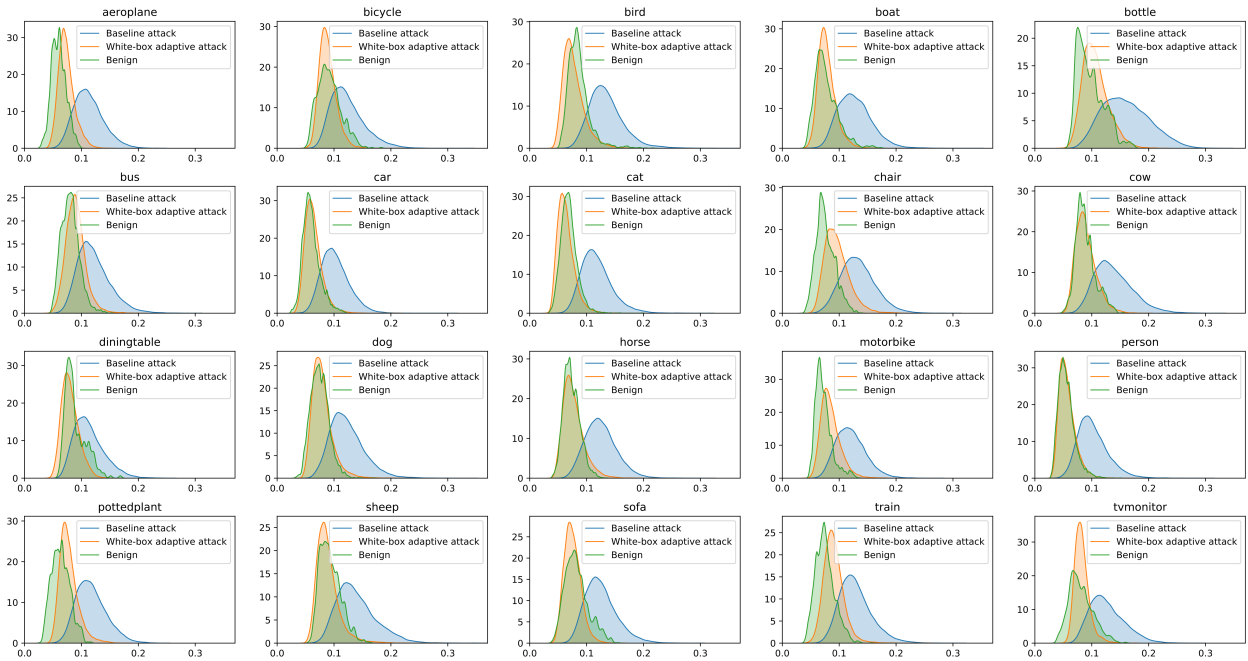}}
\caption{Reconstruction error distribution plot for each category.}
\label{fig:density_per_class}
\end{figure*}

\section{Gray-box attack performance on MS COCO dataset}
We present in the main paper the gray-box attack performance on the PASCAL VOC dataset. For completeness, in this section, we show the gray-box attack performance on the MS COCO dataset.
The surrogate system is train on \textit{coco14valminusminival}.
As shown in Table~\ref{tab:gbcoco}, gray-box attack the MS COCO dataset achieves very similar results compared to white-box attack, which is aligned with what we observe on the PASCAL VOC dataset.

\section{Ablation study on MS COCO dataset}
We show in the main paper that by constraining the perturbed area to the target object region,
the attack performance is worse, which implies that context-aware attack need to perturb not only the target region, but also other regions to achieve context consistency. In this section, we present a same ablation study on the MS COCO dataset. The results are shown in Table~\ref{tab:coco}. Similar to the results on the PASCAL VOC dataset, hiding attack is not affected by much, implying hiding attacks do not need to perturbations over other regions; however, mis-categorization and appearing attack performance is lower when the other regions are not perturbed.

 \begin{table*}
 \centering
 \caption{Gray-box attack performance for the three attack goals on the the MS COCO dataset.}
 \label{tab:gbcoco}
 \small
 \resizebox{14cm}{!}{
 \begin{tabular}{c|ccc|ccc}
 \hline
 \multirow{2}{*}{Threat Model} & \multicolumn{3}{c|}{Fooling Rate}              & \multicolumn{3}{c}{Bypass Rate}                \\ \cline{2-7} 
                                 & Mis-categorization & Hiding      & Appearing   & Mis-categorization & Hiding      & Appearing   \\ \hline
 White-Box        & 86.90\%            & 90.82\%     & 66.36\%     & 83.67\%        & 75.52\%         & 86.08\% \\ \hline

 Gray-Box         & 76.25\%            & 90.82\%     & 60.51\%     & 83.86\%        & 75.89\%         & 85.98\% \\ \hline

 \end{tabular}}
 \end{table*}

\begin{table*}[]
\centering
\caption{Attack performance when only attacking the target object region for the three attack goals on the MS COCO dataset.}

\label{tab:coco}
\small
\resizebox{14cm}{!}{
\begin{tabular}{c|ccc|ccc}
\hline
\multirow{2}{*}{Attack Region}          & \multicolumn{3}{c|}{Fooling Rate}              & \multicolumn{3}{c}{Bypass Rate}                \\ \cline{2-7}
                                        & Mis-categorization & Hiding      & Appearing   & Mis-categorization & Hiding      & Appearing   \\ \hline
Whole image       & 86.90\%            & 90.82\%     & 66.36\%     & 83.67\%        & 75.52\%         & 86.08\% \\ \hline
Target object         & 67.22\%            & 90.82\%     & 43.14\%     & 88.04\%        & 80.93\%         & 89.48\% \\ \hline
\end{tabular}}
\end{table*}

\end{document}